\documentclass[sigconf,nonacm]{acmart}

\setcopyright{none}
\acmDOI{}\acmISBN{}\acmBooktitle{}
\acmYear{2025}\copyrightyear{2025}

\newcommand{\HCAIVenue}{HCAI Workshop @ CIKM '25, November 14, 2025, Seoul, Republic of Korea}

\title{Fr\'echet Oral Distance: A Coordinate-Free Metric and Benchmark for Speech-Driven Oral Animation}

\author{Jaehong Myung}
\authornote{Lead author.}
\email{mjhsin@sogang.ac.kr}
\affiliation{
  \department{Department of Artificial Intelligence}
  \institution{Sogang University}
  \city{Seoul}
  \country{South Korea}
}

\author{Seungho Eum}
\email{seunghoeum@sogang.ac.kr}
\affiliation{
  \department{Department of Computer Science}
  \institution{Sogang University}
  \city{Seoul}
  \country{South Korea}
}

\author{Hyunjung Chung}
\email{phone1263@sogang.ac.kr}
\affiliation{
  \department{Department of Computer Science}
  \institution{Sogang University}
  \city{Seoul}
  \country{South Korea}
}

\author{Unsang Park}
\authornote{Corresponding author.}
\email{unsangpark@sogang.ac.kr}
\affiliation{
  \department{Department of Computer Science}
  \department{Department of Artificial Intelligence}
  \institution{Sogang University}
  \city{Seoul}
  \country{South Korea}
}

\renewcommand{\shortauthors}{Myung et al.}

\usepackage{amsmath}
\usepackage{multirow}
\usepackage{xspace}
\usepackage{siunitx}
\usepackage{enumitem}

\usepackage{graphicx}
\usepackage{booktabs}
\usepackage{subcaption}
\usepackage{placeins} % auto \FloatBarrier at each section
\usepackage{dblfloatfix}
\usepackage{balance}

\makeatletter
\def\fps@figure{!tbp}
\def\fps@table{!tbp}
\makeatother

\newcommand{\TIMITVis}{TIMIT-Vis\xspace}

\usepackage{fancyhdr}

\makeatletter
\fancypagestyle{firstpagestyle}{
  \fancyhf{}%
  \fancyfoot[C]{\footnotesize © 2025 Copyright held by the owner/author(s).}%
}
\makeatother

\usepackage{fancyhdr}
\fancypagestyle{HCAI@first}{%
  \fancyhf{}%
  \fancyfoot[C]{\thepage}% 원치 않으면 이 줄을 주석 처리
}

\fancypagestyle{HCAI@rest}{%
  \fancyhf{}%
  \fancyhead[LE]{\scriptsize \HCAIVenue}%
  \fancyhead[RE]{\scriptsize \shortauthors}%
  \fancyhead[LO]{\scriptsize \nouppercase{\shorttitle}}% 없으면 \@title로 대체
  \fancyhead[RO]{\scriptsize \HCAIVenue}%
  \fancyfoot[C]{\thepage}%
}

\begin{document}

\begin{abstract}
Recently, speech-driven facial and oral animation technologies have been used in various fields such as education, speech therapy, and healthcare. But methods to objectively evaluate the quality of generated results are still limited. Accordingly, we propose the Fréchet Oral Distance (FOD), which uses embeddings from an Oral Motion Encoder (OME) and a TIMIT-Vis reference distribution. In experiments, FOD reproduced model rankings similar to the coordinate-based metric, Lip Vertex Error (LVE), and showed trends consistent with user preference. Our framework provides objective and reproducible benchmarking without relying on expensive equipment or subjective evaluation, and it can be extended to compare a variety of 2D/3D models.

\end{abstract}

\maketitle

% --- Copyright notice (ACM style; UNNUMBERED footnote, no '0') ---
\begingroup
\renewcommand\thefootnote{}% 번호 표시 비움
\footnotetext{\footnotesize © 2025 Copyright held by the owner/author(s).}
\addtocounter{footnote}{-1}% 카운터 원상복구
\endgroup

% --- Page styles: page 1 clean (no header), others use HCAI@rest ---
\thispagestyle{HCAI@first}
\pagestyle{HCAI@rest}

% ---------------------------------------------------------------------
% Sections
% ---------------------------------------------------------------------
\section{Introduction}

Speech-driven facial animation generation technologies have recently been used in fields such as education and healthcare~\cite{edu1,rt-mri}. When used for language therapy such as pronunciation correction, it is important to represent oral motion naturally and accurately.

While methods for modeling facial animation continue to advance, quantitative evaluation of generated videos remains insufficient. Current evaluations fall into coordinate-based, subjective, and distribution-based approaches. Coordinate-based metrics (e.g., LVE~\cite{codetalker,meshtalk,fan2022faceformer}) require precise equipment or manual landmarks, which are hard to obtain. Subjective evaluation (e.g., MOS~\cite{mospaper}) is time- and cost-intensive and has low reproducibility. As a result, there is a lack of methods to measure the quality of oral animation accurately and reproducibly.

% 주석
\iffalse
Data limitations are also a constraint. RT-MRI~\cite{rt-mri} is small in scale and lacks speaker diversity, while VOCA~\cite{voca} and BIWI~\cite{biwi} have the advantage of precise synchronization but lack internal oral information.
\fi

To address these issues, this study proposes the following contributions:
\begin{itemize}[leftmargin=*, labelsep=0.5em]
    \item \textbf{Fr\'echet Oral Distance (FOD):} a metric that evaluates quality without coordinate GT by converting generated oral animation into latent vectors and comparing distances between distributions.
    \item \textbf{TIMIT-Vis:} a large-scale articulatory simulation dataset that visualizes TIMIT with the Vocal Tract Lab (VTL)~\cite{timit,vtl}, providing a standardized evaluation environment with diverse speakers and dialects. We use \textit{TIMIT-Vis (test)} as the reference distribution because it offers stable, audio--video-synchronized articulatory coverage across diverse speakers and dialects.
\end{itemize}

\section{Related Work}

% 1) Metrics (LVE → PSNR/SSIM → MOS)
Among evaluation methods for generated video, the widely used coordinate-based metric LVE~\cite{codetalker} requires GT coordinates or equipment-based tracking, making data collection difficult. Peak Signal-to-Noise Ratio (PSNR) and Structural Similarity Index Measure (SSIM)~\cite{wang2004image} evaluate frame-wise distortion and are useful indicators of restoration quality, but they have limitations in assessing temporal information or the naturalness of motion. Subjective evaluation such as MOS (Mean Opinion Score)~\cite{mospaper} has the advantage of reflecting human perception, yet it is costly and has low reproducibility.

% 2) Distribution-based metrics (FID → FVD → FAD/FMD/FPD)
In distribution-based evaluation of generated video, Fr\'echet-based evaluation methods are widely used. Fr\'echet Inception Distance (FID)~\cite{fid} approximates the embedding distributions extracted from Inception-V3~\cite{inception} as multivariate Gaussians and computes the Fr\'echet distance. Fr\'echet Video Distance (FVD)~\cite{fvd} extends FID to video by computing the distance between two distributions using I3D embeddings~\cite{carreira2017quo}. Domain-specific Fr\'echet-based metrics have also been proposed. For example, in audio, Fr\'echet Audio Distance (FAD)~\cite{fad} uses VGGish embeddings~\cite{vggish}; for human motion sequences, Fr\'echet Motion Distance (FMD)~\cite{maiorca2022evaluating} applies Fr\'echet computation to motion-latent embeddings; and for 3D point clouds, Fr\'echet Pointcloud Distance (FPD)~\cite{shu20193d} was proposed in the context of TreeGAN. In this way, encoders suitable for each domain are used to evaluate at the distribution level rather than via frame-wise distortion.

% fig1
\begin{figure*}[!t]
  \centering
  \includegraphics[width=\textwidth]{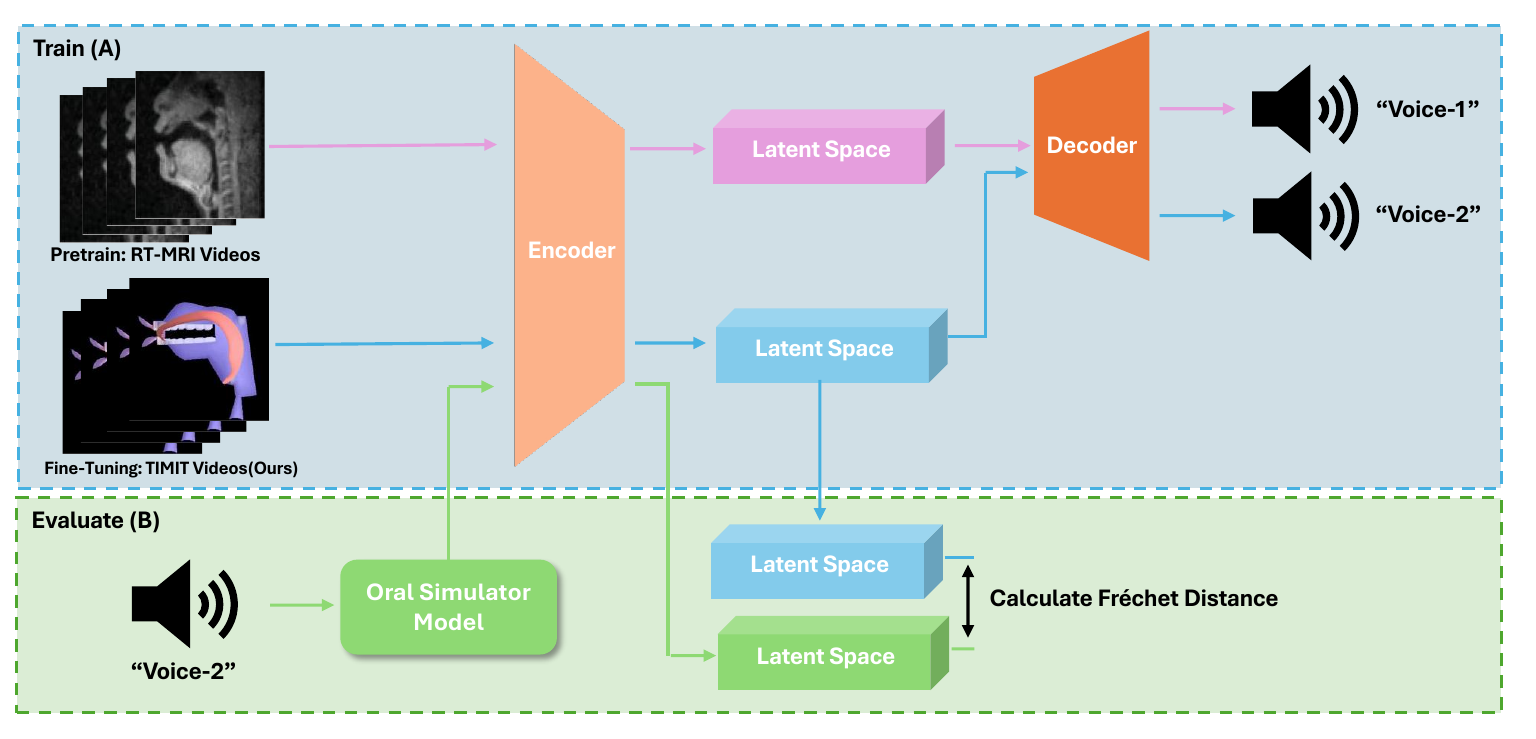}
  \caption{(A) Training pipeline: The Oral Motion Encoder is trained using synthetic articulatory videos generated from \textit{TIMIT-Vis}~\cite{timit,vtl} (VTL-rendered TIMIT). (B) Evaluation pipeline: Latent vectors extracted from simulation videos are compared to reference distributions using the Fr\'echet distance~\cite{fid}. These two stages collectively define the framework for both learning and assessing articulatory dynamics.}
  \label{fig:train}
\end{figure*}

% 3) Latent L1/L2 vs Fréchet, FOD 제안
Latent-space L1/L2 measure average deviations between embeddings. By contrast, Fr\'echet-based metrics (e.g., FID~\cite{fid}, FVD~\cite{fvd}) compare distributions by matching both means and covariances, capturing channel correlations, scale differences, and variability. Building on this perspective, our Fr\'echet Oral Distance (FOD) is a distribution-based approach configured for the domain of oral animation.

% 4) Datasets (RT-MRI → VOCA/BIWI → TIMIT-Vis 필요성)
To apply and benchmark such distribution-based metrics in our domain, a suitable articulatory dataset is necessary. RT-MRI~\cite{rt-mri} allows direct observation of internal articulators (e.g., tongue, velum) with temporal coupling to speech, yet publicly available datasets remain limited in scale and scope. VOCA~\cite{voca} and BIWI~\cite{biwi} are 4D face datasets paired with audio and offer precise topological synchronization, but they lack internal oral information. This motivates the use of TIMIT-Vis, which visualizes TIMIT with the Vocal Tract Lab (VTL)~\cite{timit,vtl}, providing audio–video-synchronized articulatory coverage across diverse speakers and dialects. These limitations restrict such datasets utility as training data for encoders that require internal oral-motion information.

% ===================== Method =====================
\section{Method}
\label{sec:method}

We propose a Fr\'echet-based metric that compares the latent distributions of generated facial and oral animations. Generated videos are embedded with an Oral Motion Encoder (OME), and detailed evaluation is defined in \S3.2.

\subsection{OME \texorpdfstring{$\times$}{x} TIMIT-Vis}
% 짧은 버전
OME encodes oral motion from generated animations into a latent space using Lip2Wav’s spatiotemporal encoder~\cite{lip2wav}. It is pretrained on RT-MRI~\cite{rt-mri} (resampled 83$\to$30~fps) and fine-tuned on \TIMITVis~\cite{timit,vtl} with the same configuration. Each input covers about 1.5~s ($\approx$45 frames) in a sliding window. The overall training process is illustrated in Figure~\ref{fig:train}A.

\begin{figure}
  \centering
  \includegraphics[width=0.72\linewidth]{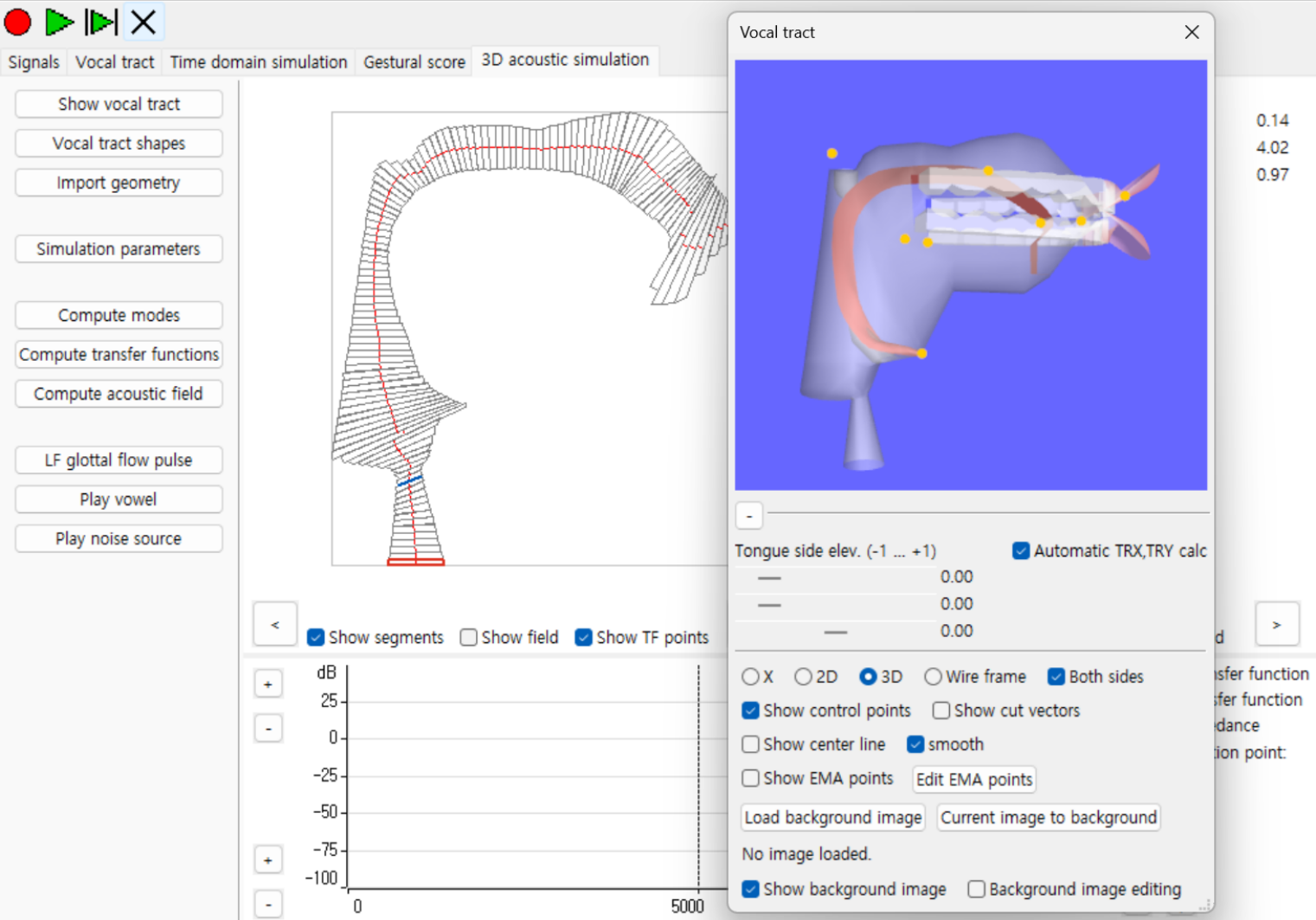}% 작게(열 너비의 약 72%)
  \caption{VocalTractLab (VTL)~\cite{vtl} interface. The GUI visualizes articulatory structures and computes
acoustic responses; VTL was used to render videos for \textit{TIMIT-Vis}~\cite{timit,vtl}.}
  \label{fig:vtl}
\end{figure}

\TIMITVis\ visualizes TIMIT~\cite{timit} speech via VTL~\cite{vtl}, rendering articulator motion at 30~fps and $256{\times}256$. It includes 6{,}200/100 train/test videos from 630 speakers (8 dialects; 192~F / 438~M) (Table~\ref{tab:timit-vis}). The dataset is used for OME training and to build the reference latent distribution. VTL was modified to cover English/German IPA, and its rendering interface is shown in Figure~\ref{fig:vtl}.

% 긴 버전
% OME maps oral motion in generated animations into a latent space. It adopts Lip2Wav’s spatiotemporal encoder~\cite{lip2wav} to learn continuous, non-linear lip-motion patterns. Training performs pretraining on RT-MRI~\cite{rt-mri} resampled \emph{from 83 fps to 30 fps} and then fine-tunes on TIMIT-Vis~\cite{timit,vtl} with the same configuration. The input uses a fixed window of about 1.5 seconds ($\approx$45 frames) applied in a sliding manner. The full training pipeline is summarized in Figure~\ref{fig:train}A.
% TIMIT-Vis is a benchmark that visualizes TIMIT~\cite{timit} speech using VTL~\cite{vtl}, rendering the motion of articulators such as the tongue, lips, and pharynx at 30~fps and $256{\times}256$. The dataset consists of Train/Test = 6,200/100, 630 speakers (8 dialect regions; 192F/438M) (see Table~\ref{tab:timit-vis}). It is used for OME training and to compute the reference latent distribution during evaluation. We also adjusted VTL to cover the International Phonetic Alphabet (IPA) for English/German to include a wider phoneme set. As an example of the rendering environment, the VTL interface used for generating \TIMITVis\ is shown in Figure.~\ref{fig:vtl}.

% ---- Table 1: Summary of the TIMIT-Vis benchmark dataset
\begin{table}[!htbp]
  \centering
  \caption{Summary of the TIMIT-Vis benchmark dataset~\cite{timit,vtl}. 6,200/100 videos (train/test) at 30 FPS, 256×256; 630 speakers, 8 dialect regions (192 female, 438 male).}
  \label{tab:timit-vis}
  \small
  \setlength{\tabcolsep}{5pt}\renewcommand{\arraystretch}{1.15}
  \begin{tabular}{|c|c|c|c|}
    \hline
    \textbf{Name} & TIMIT-Vis & \textbf{FPS} & 30 \\
    \hline
    \textbf{Resolution} & $256{\times}256$ & \textbf{Regions} & 8 \\
    \hline
    \textbf{Samples (Train/Test)} & 6,200/100 & \textbf{F/M} & 192/438 \\
    \hline
  \end{tabular}
\end{table}

% \vspace{-4pt}

% \begin{table}[h]
% \centering
% \begin{tabular}{l|l}
% \hline
% \textbf{Attribute} & \textbf{Value} \\
% \hline
% Name & TIMIT-Vis \\
% FPS & 30 \\
% Resolution & 256×256 \\
% Regions & 8 \\
% Samples (Train/Test) & 6,200 / 100 \\
% F/M & 192 / 438 \\
% \hline
% \end{tabular}
% \caption{TIMIT-Vis dataset summary (vertical format)}
% \end{table}

% ===== Figure2 : VTL interface (small) =====

% ===================== Experiments =====================
% ---- Table 3: absolute scores (leaderboard)
\begin{table*}[t]
  \caption{Absolute scores on VOCA/BIWI: LVE (when available) and latent-space L1/L2/FOD. Lower indicates better alignment.}
  \label{tab:lve_metrics}
  \centering
  {\footnotesize
    \renewcommand{\arraystretch}{1.2}
    % ▼ 마지막 열(c) 제거
    \begin{tabular}{|c|c|cc|cc|cc|cc|}
      \hline
      \multicolumn{2}{|c|}{\textbf{Metric}} & 
      \multicolumn{2}{c|}{\textbf{FaceFormer~\cite{fan2022faceformer}}} & 
      \multicolumn{2}{c|}{\textbf{CodeTalker~\cite{codetalker}}} & 
      \multicolumn{2}{c|}{\textbf{SelfTalk~\cite{peng2023selftalk}}} & 
      \multicolumn{2}{c|}{\textbf{UniTalker~\cite{fan2024unitalker}}} \\
      % ▼ 범위는 그대로 3–10 (오른쪽 열 삭제됨)
      \cline{3-10}
      \multicolumn{2}{|c|}{} & VOCA~\cite{voca} & BIWI~\cite{biwi} & VOCA~\cite{voca} & BIWI~\cite{biwi} & VOCA~\cite{voca} & BIWI~\cite{biwi} & VOCA~\cite{voca} & BIWI~\cite{biwi} \\
      \hline
      % ▼ 마지막 열 값/대시 제거
      \multicolumn{2}{|c|}{\textbf{LVE}} & 4.98 & 1.17 & 4.79 & 1.12 & 4.25 & 0.96 & 3.86 & 0.81 \\
      \hline
      \multirow{3}{*}{\textbf{Ours}} 
      & L1         & 0.0502 & 0.0542 & 0.0501 & 0.0545 & 0.0494 & 0.0532 & 0.0491 & 0.0526 \\
      & L2         & 13.85  & 14.92  & 13.77  & 14.96  & 13.61  & 14.90  & 13.42  & 14.23 \\
      & Frechet-D  & 190.10 & 165.14 & 187.36 & 167.92 & 181.51 & 163.90 & 176.94 & 162.14 \\
      \hline
    \end{tabular}
    }
\end{table*}

% ---- Figure 3: trend summary (complements Table 3)
\begin{figure*}[!t]
  \centering
  \includegraphics[width=\textwidth]{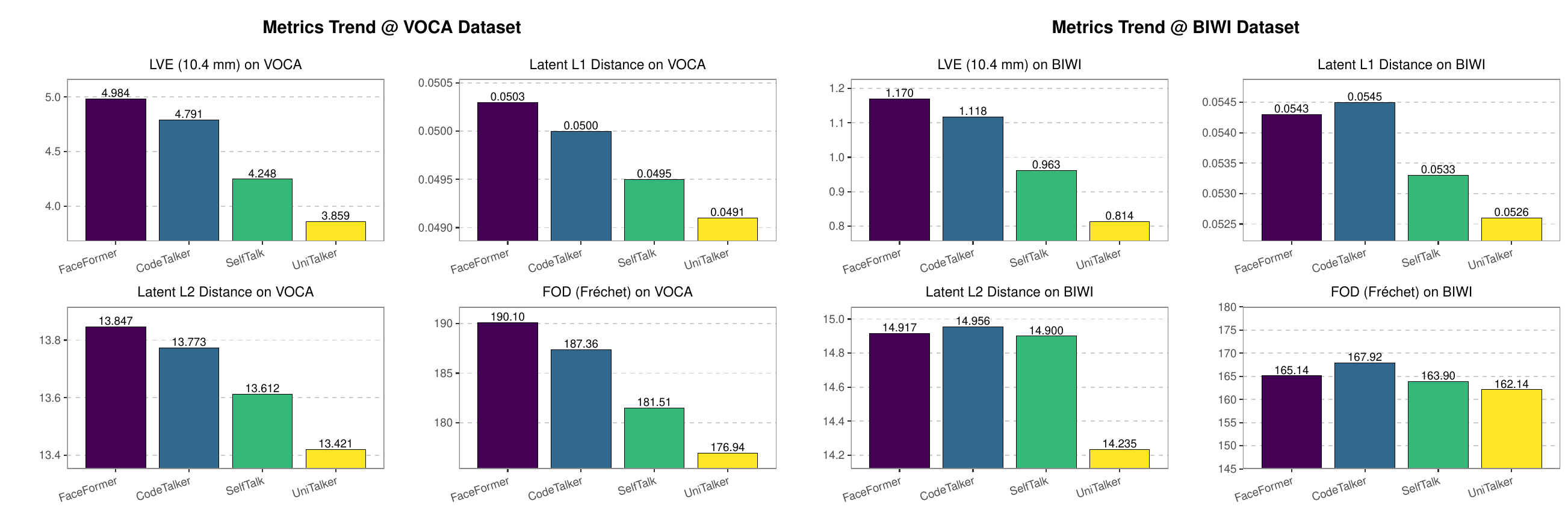}
  \caption{Trend visualization across datasets: LVE, latent L1/L2 distances, and FOD on VOCA and BIWI~\cite{voca,biwi}. The figure \textbf{summarizes ranking and separability trends} (lower is better), complementing the absolute values in Table~\ref{tab:lve_metrics}.}
  \label{fig:metrics}
\end{figure*}

\subsection{Fr\'echet Oral Distance (FOD)}
FOD is the Fréchet distance~\cite{fid} between two distributions: OME latents from generated videos and the TIMIT-Vis (test)~\cite{timit,vtl} reference. We estimate means and covariances for both and then compute the distance. Lower values indicate better alignment with the reference articulatory distribution.

%
%\begin{equation}
%\mathrm{FOD}^2
%= \left\| \hat{\boldsymbol\mu}_g - \hat{\boldsymbol\mu}_r \right\|_2^2
%+ \operatorname{tr}\!\Big(
%\tilde{\boldsymbol\Sigma}_g + \tilde{\boldsymbol\Sigma}_r
%- 2\,\big(\tilde{\boldsymbol\Sigma}_r^{1/2}\,\tilde{\boldsymbol\Sigma}_g\,\tilde{\boldsymbol\Sigma}_r^{1/2}\big)^{1/2}
%\Big),
%\quad \tilde{\boldsymbol\Sigma}_\bullet=\hat{\boldsymbol\Sigma}_\bullet+\varepsilon \mathbf{I},
%\label{eq:fod}
%\end{equation}
%

\noindent\textbf{Evaluation procedure.}
\begin{enumerate}[label=\textbf{Step~\arabic*:}, leftmargin=*, itemsep=0pt]
\item Build the reference once using \textit{TIMIT-Vis (test)}: extract OME representations and record their summary statistics (mean and covariance).
\item For each method, generate evaluation videos, extract OME embeddings from its outputs, and compute the method’s summary statistics (mean and covariance).
\item Compare the method’s summary to the fixed reference with the Fréchet Oral Distance and report the score (lower is better).
\end{enumerate}
\noindent Quantitative results are reported in a leaderboard format. These steps correspond to the evaluation pipeline depicted in Fig.~\ref{fig:train}B.

\section{Experiments}
\subsection{Experimental Setup}

\noindent\textbf{Data \& Reference.}\enspace
All methods generate rendered videos from the same 100 test audios. We embed the videos with OME and compare the latents to the fixed \textit{TIMIT-Vis (test)}~\cite{timit,vtl} reference. Baselines use publicly released VOCA~\cite{voca}/BIWI~\cite{biwi}-pretrained weights and are evaluated in batch with the same pipeline.

\noindent\textbf{Encoder Training.}\enspace
The OME is pretrained on RT-MRI (from 83 fps to 30 fps) and then fine-tuned on \textit{TIMIT-Vis}-train with the same input configuration (about 1.5 seconds; $\approx$45 frames)~\cite{rt-mri,timit,vtl}. To reduce accumulated errors on long sequences, the input length is limited to about 1.5 seconds, and audio--video pairs are randomly shuffled.

\noindent\textbf{System Configuration.}\enspace
The system environment used for training and evaluation is summarized in Table~\ref{tab:system_config}.

\begin{table}[!htbp]
  \caption{Experimental system configuration for training and evaluation.}
  \label{tab:system_config}
  \centering
  \begin{tabular}{|c|c|c|c|}
    \hline
    \textbf{OS} & Ubuntu 22.04 & \textbf{Python} & 3.10.12 \\
    \hline
    \textbf{CPU} & Xeon Gold 6248R & \textbf{Torch} & 2.0.1 \\
    \hline
    \textbf{GPU} & RTX-A6000 & \textbf{CUDA} & 12.1 \\
    \hline
  \end{tabular}
\end{table}

\noindent\textbf{Baselines.}\enspace
We include FaceFormer~\cite{fan2022faceformer}, CodeTalker~\cite{codetalker}, SelfTalk~\cite{peng2023selftalk}, and UniTalker~\cite{fan2024unitalker} using public pretrained models. All methods are tested on 100 unused audios from TIMIT-Vis, and evaluation is conducted with the same pipeline.

\noindent\textbf{Metrics.}\enspace
We report FOD as the primary metric and additionally present \emph{latent L1/L2 distances} and LVE. In this paper, latent L1/L2 distances are used as simple baselines that measure the average deviation between embeddings, whereas FOD is a distributional distance that compares the mean and covariance of OME representations against the fixed TIMIT-Vis (test)~\cite{timit,vtl} reference distribution. When the distribution is relatively simple and visibility is good, L1/L2 can also be effective; when channel correlations, scale, or variance structure are present, FOD tends to provide a more robust separation.

% \subsection{Results \& Discussion}
\subsection{Results}

\noindent\textbf{Main Results.}\enspace
Evaluating VOCA~\cite{voca} and BIWI~\cite{biwi} together, we find that on VOCA the FOD—computed as the Fréchet distance between each model and the fixed TIMIT-Vis (test)~\cite{timit,vtl} reference statistics—reproduces nearly the same ranking as LVE. The auxiliary baselines (latent L1/L2 distances) show similar trends. In contrast, on BIWI the agreement between FOD and LVE is weaker and the separability is smaller. We attribute this to BIWI’s side-view rendering, which lowers the visibility of lip contours and the lips, making it difficult for the OME to stably extract speech-related lip motion. As a result, on BIWI, FOD does not align with LVE to the same extent, with several mid-tier models swapping adjacent order (see Fig.~\ref{fig:metrics} for interpretation and Table~\ref{tab:lve_metrics} for the numbers).

\begin{figure}[!htbp]
  \centering
  \includegraphics[width=\columnwidth]{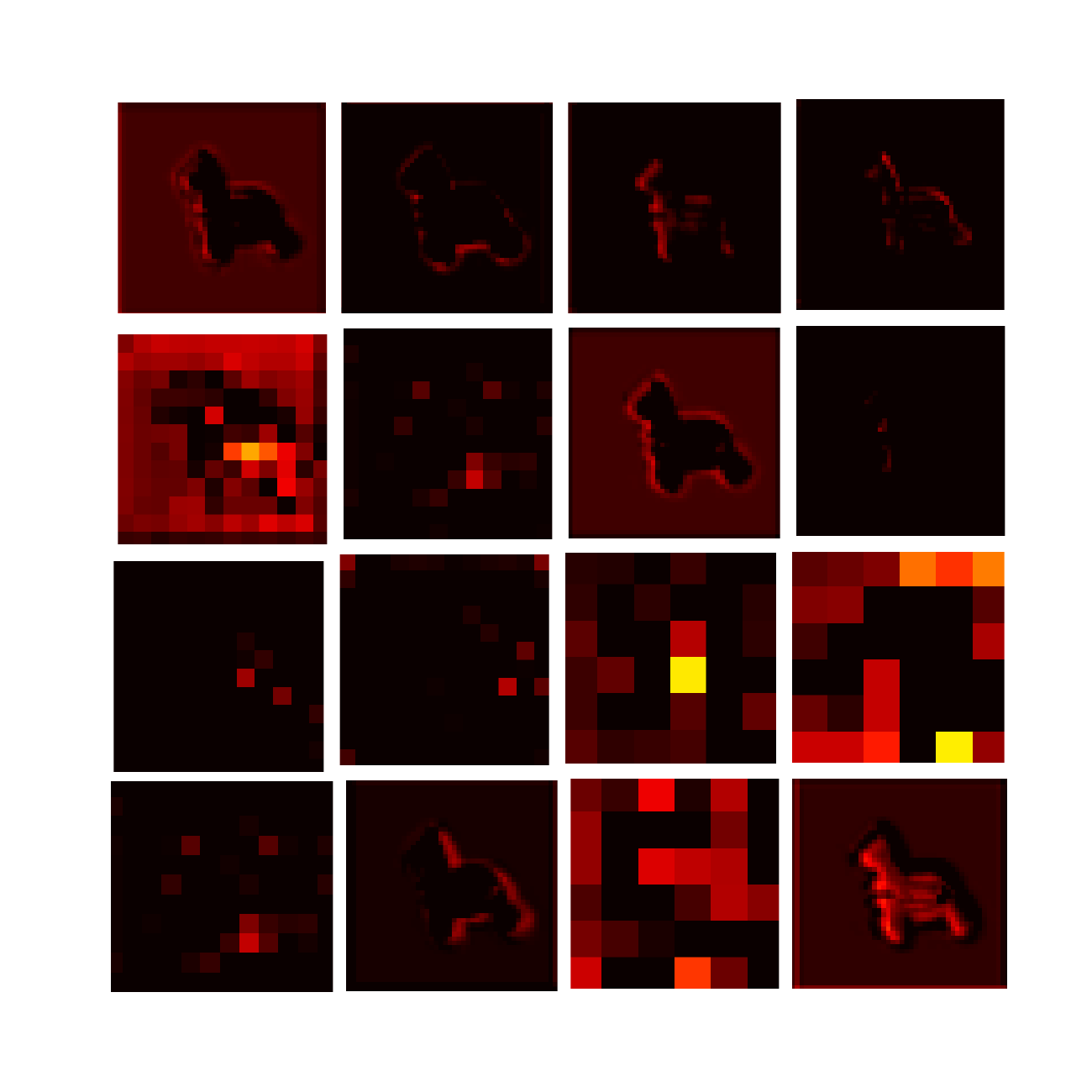}
  \caption{Activation maps of 16 latent channels (reshaped to $H\times W$). Responses highlight both articulatory motion (e.g., lips, tongue) and static oral structure, indicating joint encoding of dynamics and shape.}
  \label{fig:activation}
\end{figure}

\noindent\textbf{Qualitative Analysis.}\enspace
Some OME latent channels respond to macroscopic motions(e.g., lip opening/closing and anterior–posterior lip protrusion), whereas others are sensitive to microscopic timing. Several channels directly reflect oral contours or static shape cues, indicating that the encoder captures not only articulatory motion but also structural characteristics. As shown in the activation maps in Fig.~\ref{fig:activation}, these patterns are clearly observed.

% figure 4

% table 4
\begin{table}[!htbp]
  \centering
  \setlength\tabcolsep{3pt}
  \caption{User study (from \textit{CodeTalker}~\cite{codetalker}) on BIWI-Test-B and VOCA-Test using A/B testing. Values are percentages.}
  \label{tab:user_study}
  \resizebox{\columnwidth}{!}{%
  \begin{tabular}{lcccc}
    \toprule[0.8pt]
    \multirow{2}{*}{Competitors} & \multicolumn{2}{c}{BIWI-Test-B} & \multicolumn{2}{c}{VOCA-Test} \\
    \cmidrule(lr){2-3} \cmidrule(lr){4-5}
                                 & Lip Sync & Realism              & Lip Sync & Realism \\
    \\[-2.5ex]\hline\\[-2.2ex]
    CodeTalker \emph{vs.} VOCA       & 92.47 & 89.25 & 86.02 & 84.95 \\
    CodeTalker \emph{vs.} MeshTalk   & 80.65 & 82.80 & 95.70 & 92.47 \\
    CodeTalker \emph{vs.} FaceFormer & 53.76 & 56.99 & 70.97 & 69.89 \\
    CodeTalker \emph{vs.} GT         & 43.01 & 49.46 & 43.01 & 43.01 \\
    \bottomrule[0.8pt]
  \end{tabular}}
\end{table}

\noindent\textbf{Human Evaluation Results.}\enspace
Participants generally preferred models with lower LVE. The same ordering appears with FOD, indicating qualitative agreement with coordinate-based and perceptual judgments~\cite{codetalker}. \textit{The detailed A/B outcomes are summarized in Table~\ref{tab:user_study}.}
While FOD tracks LVE on VOCA, we stop short of making perceptual claims; direct user studies tailored to FOD are deferred to future work.

\section{Discussion}

\subsection{Limitations}
Our study has limits. Ablations of the OME are missing, so impacts of convolution depth, self-attention, and latent size on FOD are unknown. FOD assumes Gaussian latents; sensitivity to higher-order moments or KDE is unverified. Latents also encode static shape cues, implying possible shape-similarity effects on FOD. Evaluation is confined to English TIMIT-based data, leaving tonal languages and special styles (rate changes, whispering, singing) open. The fixed reference from TIMIT-Vis (test), while disjoint from OME training, makes FOD corpus-specific.

\subsection{Future Directions}
Next steps include architecture/training ablations to quantify FOD stability and discriminability; exploring higher-order statistics, kernel bandwidths, and alternative two-sample distances (e.g., MMD) in Fr\'echet~\cite{fid} computations; and enforcing disentanglement (synthetic controls, channel masking, orthogonality) to separate motion from shape. We aim to extend \textit{TIMIT-Vis}~\cite{timit,vtl} to multilingual, multi-speaker and clinical use, and to run user studies~\cite{mospaper}. View-invariant or multi-view processing could reduce BIWI side-view sensitivity.

\section{Conclusion}
We presented a framework for evaluating speech-driven facial animation.
It combines a distribution-based metric (Fréchet Oral Distance, FOD) with TIMIT-Vis~\cite{timit,vtl}, an articulatory simulation dataset with guaranteed audio--video synchronization.
Together, they mitigate reliance on coordinate GT, reduce the instability of subjective evaluation, and address data-diversity gaps.
In our experiments, FOD reproduced rankings aligned with LVE and showed higher discriminability than latent L1/L2 distances, and it also qualitatively agreed with user preference.
In addition, channel activation visualizations support that OME encodes both motion and structural characteristics.
TIMIT-Vis provides a stable reference distribution with diverse speakers and dialects and is meaningful as a standard public resource for future evaluation research.
The proposed framework functions as an objective and reproducible benchmarking tool and offers extensibility to compare various 2D/3D generative models under a common standard.

% ---------------------------------------------------------------------
% Acknowledgments
% ---------------------------------------------------------------------
\begin{acks}
% \section*{Acknowledgments}
This work was supported by Institute of Information \& communications Technology Planning \& Evaluation (IITP) grant funded by the Korea government(MSIT) (No. 2022-0-00621, RS-2022- II220621, Development of artificial intelligence technology that provides dialog-based multi-modal explainability)
\end{acks}

% ---------------------------------------------------------------------
% References
% ---------------------------------------------------------------------
\FloatBarrier
\clearpage
\balance
\bibliographystyle{ACM-Reference-Format}
\bibliography{main}

\end{document}